\documentclass[a4paper,french]{rnti}
\usepackage{graphicx}
\usepackage{amssymb, amsthm, amsmath, amsfonts}
\usepackage{subfigure,graphicx}
\usepackage[utf8]{inputenc}
\usepackage[T1]{fontenc}

\DeclareMathOperator{\ocap}{\displaystyle{\small{\textcircled{{\scriptsize $\cap$}}}}}

\titrecourt{Une mesure d'expertise pour le {\em crowdsourcing}}

\nomcourt{H. Ouni et al.}

\titre{Une mesure d'expertise pour le {\em crowdsourcing}}

\auteur{Hosna Ouni\affil{1},
		Arnaud Martin\affil{1}, 
		Laetitia Gros\affil{2},
		Mouloud Kharoune\affil{1},
		Zoltan Miklos\affil{1}}

\affiliation{
   \affil{1}UMR 6074 IRISA, DRUID team, Université de Rennes 1, Lannion, France\\
          hosnaouni@gmail.com, \{arnaud.martin, mouloud.kharoune,zoltan.miklos\}@univ-rennes1.fr,\\
          \http{http://www-druid.irisa.fr}\\
   \affil{2} Orange Labs, 2, av. Pierre Marzin, F-22307 Lannion Cedex, France \\
          laetitia.gros@orange.com\\ }

\resume{%
Le {\em crowdsourcing}, un enjeu économique majeur, est le fait d'externaliser une tâche interne d'une entreprise vers le grand-public, la foule. C'est ainsi une forme de sous-traitance digitale destinée à toute personne susceptible de pouvoir réaliser la tâche demandée généralement rapide et non automatisable. L'évaluation de la qualité du travail des participants est cependant un problème majeur en {\em crowdsourcing}. En effet, les contributions doivent être contrôlées pour assurer l'efficacité et la pertinence d'une campagne. Plusieurs méthodes ont été proposées pour évaluer le niveau d'expertise des participants. Ce travail a la particularité de proposer une méthode de calcul de degrés d'expertise en présence de données dont l'ordre de classement est connu. Les degrés d'expertise sont ensuite considérés sur des données sans ordre pré-établi. Cette méthode fondée sur la théorie des fonctions de croyance tient compte des incertitudes des réponses et est évaluée sur des données réelles d'une campagne réalisée en 2016.
}

\summary{
Crowdsourcing, a major economic issue, is the fact that the firm outsources internal task to the crowd. It is a form of digital subcontracting for the general public.
The evaluation of the participants work quality is a major issue in crowdsourcing. Indeed, contributions must be controlled to ensure the effectiveness and relevance of the campaign. We are particularly interested in small, fast and not automatable tasks. Several methods have been proposed to solve this problem, but they are applicable when the "golden truth" is not always known. This work has the particularity to propose a method for calculating the degree of expertise in the presence of gold data in crowdsourcing. This method is based on the belief function theory and proposes a structuring of data using graphs. The proposed approach will be assessed and applied to the data.
}

\begin{document}
%\layout

% DEBUT DE L'ARTICLE
%
\section{Introduction}
Le {\em crowdsourcing}, concept lancé par \cite{howe2006rise}, stimule la participation collective à l'élaboration de certaines tâches qu'une entreprise ne souhaite pas réaliser en interne par faute de ressources ou de temps et qu'il est compliqué voire impossible de confier à un ordinateur. Il s'inscrit dans une logique de partage dérivée de l'essor du web 2.0. En effet, l'échange des idées et des savoir-faire se réalise par l'intermédiaire d'une plateforme internet. 

Plusieurs plateformes telles que Amazon Mechanical Turk (AMT), Microworker et Foule Factory sont destinées aux petites tâches que la machine est incapable d'effectuer rapidement et de façon fiable. Ces tâches sont généralement simples et courtes à l'instar de l'analyse des émotions, la catégorisation des produits ou la comparaison de designs. 

Néanmoins, l'environnement d'une plateforme de {\em crowdsourcing} est incertain car finalement peu maîtrisé. Par suite, l'évaluation de la qualité et la fiabilité des contributions et des travailleurs eux-même est indispensable pour le bon déroulement d'un tel processus. Ainsi, plusieurs travaux ont été proposés pour identifier les experts ou les travailleurs sérieux de la plateforme.

Ce papier propose une solution permettant de calculer la pertinence des réponses des participants à partir des réponses à une campagne lancée par Orange Labs. Durant cette campagne, on se propose  de faire évaluer par les travailleurs de la plateforme de {\em crowdsourcing} la qualité sonore restituée par différentes solutions de codage audio. La procédure consiste à faire écouter aux participants 12 extraits musicaux de différentes qualités et à leur demander d'en évaluer la qualité audio sur une échelle à 5 catégories (Excellente, Bonne, Moyenne, Médiocre, Mauvaise). A chaque catégorie est associée une note allant de 1 (pour Mauvais) à 5 (pour Excellent) voir \cite{ITU96}. Parmi les extraits sonores présentés dans un ordre aléatoire, 5 sont de qualité connue (ajout d'un bruit modulé par le signal, avec différents rapports signal/bruit (MNRUs~: {\em Modulated Noise Reference Unit} voir \cite{ITU96})). Ces signaux sont utilisés dans les tests comme références et ancrages sur l'échelle de qualité. Dans cette étude, les signaux MNRU vont permettre de définir des degrés d'expertise. L'idée est de structurer les réponses des travailleurs par des graphes, représentant un ordre de préférences (issu des notes) entre les signaux MNRU, puis de comparer ces graphes avec celui de référence issu des notes théoriques attendus. Ensuite, cette estimation sera prise en compte pour sélectionner les participants experts pour les 7 autres extraits sonores afin d'atteindre l'objectif souhaité de classement de ces signaux ({\em i.e.} des codeurs dont ils sont issus).

La comparaison des graphes consiste à quantifier la similarité entre deux graphes. Il s'agit d'un problème très courant surtout dans les réseaux sociaux car les graphes révèlent des propriétés topologiques qu'on cherche à comprendre et à comparer. Malheureusement, on manque de références permettant de construire une échelle de comparaison de leurs caractéristiques géométriques. En effet, il n'y a pas de métrique ou de méthode de comparaison de graphes universelle. De plus, du point de vue algorithmique, les méthodes classiques pour aborder ce genre de problèmes sont complexes.

De plus, les réponses étant fournies pas des humains dans un environnement non contrôlé, contrairement aux tests d'écoute classiques en laboratoire, il est nécessaire de modéliser les imperfections sur les réponses. La théorie des fonctions de croyance permet de répondre à cette problématique et d'offrir un cadre théorique pour réaliser la combinaison des informations issues de différentes sources.

Nous proposons donc dans ce travail une approche originale permettant une estimation d'une mesure d'expertise à partir d'une comparaison de graphes dans le cadre de la théorie des fonctions de croyance. Ainsi, la section suivante présente les concepts de base de la théorie des fonctions de croyance. La section \ref{Meth}, après un bref rappel des approches existantes, propose une approche originale pour la représentation des réponses sous forme de graphes et le calcul de degrés d'expertise. Finalement, l'évaluation de la méthode sur des données réelles fait l'objet de la section \ref{result}.

\section{La théorie des fonctions de croyance}
\label{Croyance}
La théorie des fonctions de croyance issue des travaux de \cite{dempster1967upper} et de \cite{shafer1976mathematical} permet une représentation à la fois des incertitudes et des imprécisions mais aussi de l'ignorance d'une source (ici la réponse d'un contributeur). Considérant un ensemble $\Omega=\{\omega_1,\omega_2,..,\omega_n\}$ qui représente les réponses possibles à une question, une fonction de masse est définie sur $2^\Omega$ (ensemble de toutes les disjonctions de $\Omega$) et à valeur dans $[0,1]$ avec les contraintes~:
\begin{equation}
	\label{bba}
	\begin{cases}
	\sum_{A\subseteq\Omega} m(A)=1\\
	m(\emptyset)=0
	\end{cases}
\end{equation}
La fonction de masse $m(A)$ représente la part de croyance allouée à la proposition $A$ et qui ne peut pas être affectée à un sous-ensemble strict de $A$. Elle peut être vue comme une famille d'ensembles pondérées ou une distribution de probabilité généralisée. Un ensemble $A$ est un élément focal si $m(A)\neq 0$. Par exemple, si nous considérons la fonction de masse $m(\{\omega_1, \omega_2\})=0.8$, $m(\Omega)=0.2$, cette quantité représente une imprécision sur $\omega_1$ ou $\omega_2$ et une incertitude car la valeur affectée à cette proposition est 0.8.

La manipulation des données imparfaites issues de plusieurs sources distinctes nécessite de fusionner les informations. On parle donc de la combinaison des fonctions de masse permettant l'aboutissement à un état de connaissance générique et pertinent. 
L'opérateur de combinaison conjonctive proposé par \cite{smets} est donné pour deux fonctions de masse issues de deux sources par~: 
\begin{equation}
\label{comb_smets}
(m_{1}\ocap m_{2})(A)= \sum_{B_{1} \cap B_{2}=A}m_{1}(B_{1})m_{2}(B_{2})
\end{equation}
La masse affectée sur l'ensemble vide à l'issue de cette combinaison peut être interprétée comme l'inconsistance de la fusion.

Afin de mesurer l'écart à une fonction de masse attendue, par exemple pour prendre une décision ou définir une mesure, plusieurs distances ont été proposées. La distance de \cite{Jousselme} est la plus communément utilisée pour ses propriétés de répartitions des pondérations en fonction de l'imprécision des éléments focaux. Elle est donnée par~:
\begin{equation}
\label{Jouss}
d_J (m_1,m_2 )=\frac{1}{2} (m_1-m_2 )^T \underline{\underline{D}} (m_1-m_2 )
\end{equation}
avec~: 
\begin{eqnarray}
\underline{\underline{D}}(X,Y)=\begin{cases}
1 \mbox{ si } X=Y=\emptyset\\
\displaystyle \frac{\arrowvert X \cap Y\arrowvert}{\arrowvert X \cup Y\arrowvert} \,\, \forall X,Y \in 2^\Omega
\end{cases}
\end{eqnarray}

\section{Caractérisation d'experts dans le crowdsourcing}
\label{Meth}
\subsection{Positionnement du travail}

L'identification des experts sur les plateformes de crowdsourcing a fait l'objet de plusieurs travaux récents. Il faut distinguer deux types d'approches, celles tenant compte de questions dont on connaît la réponse (nommées données d'or) et celles où aucunes connaissances {\em a priori} n'est disponible. En effet, dans le contexte de ce dernier type d'approches, \cite{Am1,Am2} se sont intéressés à calculer un degré d'exactitude et de précision en supposant que la majorité a raison et en définissant ce degré à partir de la \textit{distance de} \cite{Jousselme} entre sa réponse et la moyenne des réponses des autres participants. D'autre part, \cite{DandS} et \cite{ipr} ont utilisé l'algorithme Expectation-Maximisation (E.M) permettant dans une première phase d'estimer la bonne réponse pour chaque tâche, en utilisant des étiquettes affectées par les participants, puis, d'évaluer la qualité des travailleurs en comparant les réponses soumises à la bonne réponse inférée. Également, \cite{smyth} et \cite{raykar} ont utilisé cette approche pour les classements binaires et les étiquetages catégoriques. \cite{raykar2} ont généralisé l'idée sur les classements ordinaires (associer des notes de 1 à 5 en fonction de la qualité d'un objet ou d'un service). Ces méthodes proposent de calculer la "sensibilité" (les vrais positifs) et la "spécificité" (les vrais négatifs) pour chaque label. Ainsi, le participant est un spammer si son score est proche de 0~; un expert parfait a un score égal à 1.

Cependant, ces algorithmes proposent de déterminer la qualité des réponses des participants quand la vérité est inconnue alors que dans notre cas, les notes correctes théoriques attribuées aux signaux MNRUs sont connues.  En effet, nous cherchons plutôt à identifier les experts en nous fondant sur des données correctes de référence et définir un degré d'expertise proportionnel à la similarité entre les réponses d'un participant et les réponses connues à l'avance. Ainsi, notre travail est fondé sur des données d'or qui servent à estimer directement la qualité des participants tel que proposé par \cite{Le}. Les données d'or sont des questions dont les réponses correctes sont connues à l'avance et qui sont injectées d'une façon arbitraire. Les données d'or ont l'avantage de mesurer explicitement la précision des travailleurs et d'être utilisées pour prendre des décisions concernant le travailleur~: pouvons-nous utiliser leur travail~? Faut-il les laisser finir la tâche~? Est-ce qu'ils méritent un bonus~? En second lieu, c'est processus transparent dans le sens où on s'assure que les travailleurs comprennent les détails nuancés ou difficiles des exigences de la tâche.

Pour évaluer l'impact de l'utilisation des données d'or, \cite{panos} a examiné la performance de l'algorithme de \cite{DandS} modifié pour tenir compte de l'existence de données d'or. En variant le pourcentage des données d'or (0\%, 25\%, 50\% et 75\%), il a essayé de mesurer à chaque fois l'erreur de classification (dans quelle mesure l'algorithme estime la classe correcte des exemples) et l'erreur d'estimation de la qualité (dans quelle mesure l'algorithme estime la qualité des travailleurs). Il a trouvé que ce genre de données ne fait pas de différence par rapport au modèle non supervisé. Par contre, il a considéré que l'utilisation des données d'or est nécessaire dans quelques cas tel que le travail sur des ensembles de données très déséquilibrés (évaluer toutes les classes). Selon \cite{panos}, les raisons les plus importantes sont le gain de la confiance des personnes non techniques (en proposant une approche de contrôle de qualité) et le calibrage des résultats lorsque la sensibilité des utilisateurs influent sur leurs réponses.

Également, \cite{patric} traite les données d'or comme un outil pour associer des scores de confiance aux contributeurs. Les participants doivent donc dépasser des seuils de confiance minimum pour continuer à travailler sur une tâche. Si à tout moment un contributeur tombe en dessous du seuil de confiance, on exclue son travail.

Dans ce travail, nous allons pondérer les réponses des contributeurs à partir des relations sur les réponses des données d'or. Cette pondération, réalisée à partir d'un degré d'expertise, pourra aller jusqu'à ne plus considérer les contributeurs trop éloignés des réponses attendues sur les données d'or.

\subsection{Calcul d'un degré d'expertise}
\begin{figure}[!h]
	\centering
	\caption{Graphe complet}
	\label{grcomp}
	\includegraphics[width=0.7\textwidth]{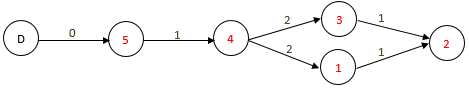}	
\end{figure}

	Comme déjà mentionné, les réponses des contributeurs de la plateforme sont représentées par des graphes. Pour cela, considérant un participant $p$ qui a associé les notes présentées dans le tableau \ref{expnote} aux MNRUs (les données d'or), le graphe orienté et pondéré correspondant est construit ({\em cf.} figure~\ref{grcomp}).
\begin{table}[!h]
	\centering
	\caption{Exemple de notes}
	\label{expnote}
	\begin{tabular}{|c|c|c|c|c|c|}
		\hline
		\textbf{MNRU}&1&2&3&4&5\\ \hline
		\textbf{Note associée}&2&1&2&4&5\\ \hline
	\end{tabular}
\end{table}
Le graphe est orienté pour la lisibilité des préférences ( $a\rightarrow b$ signifie $a$ est préféré à $b$) et la facilité du travail sur les successeurs et les prédécesseurs. Il est conçu comme suit~: 
	\begin{itemize}
		\item L'insertion du point de départ virtuel $D$, correspondant toujours à la note 5, qui sert à l'extraction de la note associée au morceau $MNRU_i$ de la façon suivante~: 
		\begin{equation}
		\label{cgnotass}
		note.ass(MNRU_i )=5-d_G(D,i)
		\end{equation}
		\item A chaque itération $k$, l'ensemble des n\oe{}uds ayant la $k^{eme}$ note la plus élevée est recherché et cet ensemble est ajouté au graphe dans une même profondeur en respectant les normes suivantes~: 
		\begin{itemize}
			\item[\textbullet] Coût de l'arc~: différence entre les notes des MNRUs de deux profondeurs consécutives.
			\item[\textbullet] Nom du n\oe{}ud~: numéro du MNRU
		\end{itemize}
	\end{itemize}

\begin{figure}[!h]
	\centering
	\caption{Graphe de référence}
	\includegraphics[width=0.5\textwidth]{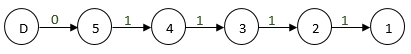}
	\label{graphcorr}
\end{figure}

Pour calculer les degrés d'expertise, nous considérons le graphe de référence qui correspond aux notes théoriques attendues pour les MNRU et données par le tableau \ref{notecorr} qui correspond au graphe \ref{graphcorr}.
\begin{table}[!h]
	\centering
	\caption{Notes correctes}
	\begin{tabular}{|c|c|c|c|c|c|}
		\hline
		\textbf{MNRU}&1&2&3&4&5\\ \hline
		\textbf{Note associée}&1&2&3&4&5\\ \hline
	\end{tabular}
	\label{notecorr}
\end{table}
Les graphes correspondants aux réponses des participants sont comparés à ce graphe de référence et une fonction de masse est ainsi calculée pour chaque réponse des contributeurs.

L'idée est ici d'extraire un ensemble de critères hétérogènes permettant de contourner les différences entre deux n{\oe}uds, ce qui est proche de la notion de "signature des n{\oe}uds" introduite par \cite{Jouili11}, bien que les critères considérés ne soient pas les mêmes.

Une particularité de la modélisation proposée est que tous les graphes ont les mêmes n{\oe}uds (même nombre et même attribut). Ainsi, l'idée est de comparer tous les couples $(N_{(1,i)},N_{(2,i)})$ où $N_{(1,i)}$ est le n{\oe}ud d'attribut $i$ dans le graphe de référence et  $N_{(2,i)}$ est le n{\oe}ud de même attribut appartenant au graphe à comparer.

Pour ce faire, nous avons défini pour chaque n{\oe}ud du graphe quatre critères représentant les différentes erreurs possibles que nous avons identifiées. Ces critères sont représentés puis fusionnés à l'aide des fonctions de masse où le cadre de discernement considéré est~: 
\begin{equation}
\label{notrecadisc}
\Omega=\{E,NE\}
\end{equation}
où $E$ représente l'assertion {\em Expert} et $NE$ {\em Non expert}. Nous cherchons ainsi à mesurer la croyance dans le fait qu'un contributeur soit un expert ({\em i.e.} détermine l'ordre correct sur les MNRUs) en fonction des notes qu'il a attribuées sur les MNRUs. \\
\textbf{1. Degré d'exactitude de la note associée}~: Ce critère est caractérisé par la différence de position d'un n{\oe}ud considéré entre la référence et la réponse du contributeur. La dissimilarité est calculée à l'aide de la distance Euclidienne dans \eqref{d1}.
	\begin{equation}
	\label{d1}
	d_{1}(N_{1,i},N_{2,i})=|d_{G_1}(D,N_{1,i})-d_{G_2}(D,N_{2,i})|
	\end{equation}
	où $d_G(D,N_{i})$ est la profondeur du n{\oe}ud $N_{i}$ par rapport au n{\oe}ud $D$.\\
	La fonction de masse correspondante à ce critère est donnée par~:
	\begin{equation}
	\label{m1}
	\begin{cases}
	m_{1}(N_{1,i},N_{2,i})(E)=\displaystyle 1-\frac{d_{1}(N_{1,i},N_{2,i})}{d_{max}}\\
	m_{1}(N_{1,i},N_{2,i})(NE)=\displaystyle \frac{d_{1}(N_{1,i},N_{2,i})}{d_{max}}
	\end{cases}
	\end{equation}\\
	où $d_{max}$ est la distance maximale entre deux n{\oe}uds. Compte tenu du fait que les graphes considérés ne représente que 5 notes, $d_{max}=4$. \\
	\textbf{2. Degré de confusion entre les MNRUs}~: Ce critère mesure la proportion des n{\oe}uds de même note/distance au point de départ $D$ que le n{\oe}ud concerné. La dissimilarité de Jaccard sera ainsi employé pour la comparaison des contenus des ensembles dans \eqref{d2} 
	\begin{equation}
	\label{d2}
	d_{2}(N_{1,i},N_{2,i})=\displaystyle \frac{|I_{N_{1,i}}\bigcap I_{N_{2,i}}|}{|I_{N_{1,i}}\bigcup I_{N_{2,i}}|} 
	\end{equation}
	où $I_{N_i}=\{N_j \in V;d_G(D,N_j)=d_G(D,N_i)\}$, avec $V$ l'ensemble des n{\oe}uds du graphe.\\
	La fonction de masse associée est donnée par~:
	\begin{equation}
	\label{m2}
	\begin{cases}
	m_{2}(N_{1,i},N_{2,i})(E)=d_{2}(N_{1,i},N_{2,i})\\
	m_{2}(N_{1,i},N_{2,i})(NE)=1-d_{2}(N_{1,i},N_{2,i})
	\end{cases}
	\end{equation}\\
	Cette masse représente une valeur minimale de 0.2.\\
	\textbf{3-4. Degré de mauvais ordre précédent} (sur l'ensemble de prédécesseurs) et \textbf{de mauvais ordre suivant} (sur l'ensemble de successeurs). Le participant peut considérer un morceau meilleur qu'un autre, contrairement à ce qui est attendu. Ainsi ces critères mesurent ces erreurs d'inversion par rapport au précédent ou suivant. Afin de définir ces degrés, nous introduisons la définition des ensembles suivants respectivement pour l'ensemble des prédécesseurs ($P_{N_i}^C$ correctes et $P_{N_i}^{NC}$ non correctes)  et l'ensemble des successeurs ($S_{N_i}^C$ correctes et $S_{N_i}^{NC}$ non correctes)~:\\
	$\begin{cases}
	P^C_{N_i}=\{N_j \in V / N_j \in  Pred_{G_1}(N_i)\}\\
	P^{NC}_{N_i}=\{N_j \in V  / N_j \in  Succ_{G_1}(N_i)\}%\\
%	P_{N_i}=$ Ensemble de prédécesseurs de $N_i
	\end{cases}$
	et 
	$\begin{cases}
	S^C_{N_i}=\{N_j \in V / N_j \in  Succ_{G_1}(N_i)\}\\
	S^{NC}_{N_i}=\{N_j \in V / N_j \in Pred_{G_1}(N_i)\}\}%\\
%	S_{N_i}= $Ensemble de successeurs de$ N_i
	\end{cases}$\\
	où $Succ_G(N)$ et $Pred_G(N)$ sont respectivement l'ensemble des successeurs et l'ensemble des prédécesseurs du n\oe{}ud $N$ dans le graphe $G$.\\
	A partir de ces définitions, les distances $d_3$ et $d_4$ sont données par les équations \eqref{d3} et \eqref{d4}.
	\begin{equation}
	\label{d3}
	\begin{cases}
	d_{3,1}(N_{1,i},N_{2,i})=\displaystyle \frac{|P^C_{N_{2,i}}\bigcap P_{N_{1,i}} |}{|P_{N_{1,i}} \bigcup P_{N_{2,i}} |}=m_{3}(N_{1,i},N_{2,i})(E)\\
	d_{3,2}(N_{1,i},N_{2,i})=\displaystyle \frac{|P^{NC}_{N_{2,i}} |}{|P_{N_{2,i}}|}=m_{3}(N_{1,i},N_{2,i})(NE)
	\end{cases}
	\end{equation}
	\begin{equation}
	\label{d4}
	\begin{cases}
	d_{4,1}(N_{1,i},N_{2,i})=\displaystyle \frac{|S^C_{N_{2,i}}\bigcap S_{N_{1,i}} |}{|S_{N_{1,i}} \bigcup S_{N_{2,i}} |}=m_{4}(N_{1,i},N_{2,i})(E)\\
	d_{4,2}(N_{1,i},N_{2,i})=\displaystyle \frac{|S^{NC}_{N_{2,i}} |}{|S_{N_{2,i}}|}=m_{4}(N_{1,i},N_{2,i})(NE)
	\end{cases}
	\end{equation}\\
	Le reste de la masse sera associé à l'ignorance. La masse associée à l'ignorance peut être également dérivée des n{\oe}uds extrêmes qui n'ont pas des prédécesseurs (tous les n{\oe}uds sauf le n{\oe}ud (5)) ou bien des successeurs (tous les n{\oe}uds sauf le n{\oe}ud (1)).

Les équations \eqref{d1}, \eqref{m1}, \eqref{d2}, \eqref{m2}, \eqref{d3} et \eqref{d4} permettent de calculer les fonctions de masse par critère pour \textbf{chaque couple de n{\oe}uds} $(N_{1,i},N_{2,i})$, respectivement du graphe de référence et du graphe correspondant à la réponse d'un participant, et d'attribut $i$. L'étape suivante définit une fonction de masse sur le \textbf{graphe tout entier} en faisant la moyenne des fonctions de masse sur tous les n{\oe}uds, calculées pour chaque critère~:
	\begin{equation}
	\label{mbba_crit}
	\begin{cases}
	m_{k}(G_{1},G_{2})(E)=\displaystyle \frac{\displaystyle \sum_{i=1}^{O(G)}m_{k}(N_{1,i},N_{2,i})(E)}{O(G)} \\
	m_{k}(G_{1},G_{2})(NE)=\displaystyle \frac{\displaystyle \sum_{i=1}^{O(G)}m_{k}(N_{1,i},N_{2,i})(NE)}{O(G)}
	\end{cases}
	\end{equation}
	où $O(G)$ est l'ordre du graphe ({\em i.e.} le nombre de sommets, ici 6).\\ 
	Afin d'obtenir une fonction de masse pour la réponse considérée, nous combinons les fonctions de masse des quatre critères.
Finalement, le degré d'expertise est donné en calculant la distance de \cite{Jousselme} entre la fonction de masse ainsi obtenue et la fonction de masse catégorique sur l'élément expert tel que \cite{Essaid2014}.

\section{Évaluation de la méthode en situation réelle}
\label{result}

Historiquement, Orange Labs réalise des tests d'évaluation subjective de codeurs audio en laboratoire. Ces tests consistent à recruter des auditeurs dits naïfs (n’étant pas impliqués directement dans les travaux liés à l'évaluation de la qualité ou du codage audio), à leur présenter de courtes séquences de parole ou de musique traitées selon différentes configurations de codage et à leur demander d’en évaluer la qualité audio sur des échelles adaptées. Les tests se déroulent dans des salles traitées acoustiquement et plus globalement dans un environnement parfaitement contrôlé.
Ces méthodes en laboratoire sont efficaces mais restent coûteuses et peuvent avoir une portée limitée quant à la représentativité des résultats (par rapport à une utilisation de services in situ) ou des stimuli (nombre limité par exemple).

Dans l'objectif d’ajouter l’approche crowdsourcing aux méthodes de test, deux campagnes déployées sur une plateforme de {\em crowdsourcing} ont été réalisées en vue de comparer les résultats avec ceux obtenus en laboratoire. Chaque campagne consistait en une réplique d’un même test initialement réalisé en laboratoire pour la normalisation du codeur G729EV.  Dans ce test laboratoire, 7 conditions de test i.e. solutions de codage étaient considérées, auxquelles s’ajoutaient les 5 conditions de référence MNRU. Au total, douze conditions étaient évaluées à travers 16 extraits musicaux. 32 personnes ont été recrutées et réparties en 4 groupes. Chaque groupe écoutait et évaluait 4 {\em hits}~, un {\em hit} ({\it Human Intelligence Task}) étant ici un ensemble de 12 séquences audio correspondant aux douze conditions de tests présentées à travers 12 extraits musicaux différents. Ainsi, chaque {\em hit} contenait les douze conditions de test (7 conditions de codage et les 5 MNRU) présentées dans un ordre aléatoire avec un set d’extraits musicaux différent pour chaque {\em hit}. Après chaque séquence audio, les auditeurs étaient invités à noter la qualité sur une échelle de 1 (= Mauvaise)  à 5 (= Excellente). 

Pour les campagnes de crowdsourcing, les participants ont également été répartis en \textbf{4 panels} de façon à ce que chacun ne puisse appartenir qu'à un et un seul panel, comme en laboratoire. Suivant le plan expérimental du test laboratoire, à chaque panel étaient associés 4 {\em hits} de 12 séquences audio à évaluer sur la même échelle de qualité. Chaque {\em hit} faisait l’objet d’un micro-job sur la plateforme de crowdsourcing. Ainsi chaque participant pouvait faire entre 1 et 4 {\em hit} (les 4 {\em hits} de son panel, différents de ceux des autres panels). La participation d’un participant a été prise en compte s’il ou elle avait terminé au moins un {\em hit}, sachant qu’il ou elle pouvait cesser les écoutes avant la fin du {\em hit} contrairement au laboratoire. Les instructions étaient présentées en anglais par écrit aux participants avant le test. Une session d’apprentissage avec 8 séquences audio devait également être réalisée avant le test, comme en laboratoire. Deux campagnes ont ainsi été menées en considérant deux zones géographiques différentes. Dans la première campagne, tout travailleur anglophone pouvait participer, quel que soit son pays. Les travailleurs ayant participé à cette campagne étaient majoritairement en Asie. La deuxième campagne se limitait aux USA. Les mêmes conditions sont appliquées dans le sens où les participants appartenant à un même $Panel_i$ écoutent les mêmes morceaux pour les deux campagnes.

\begin{figure}[!h]
		\includegraphics[width=0.9\textwidth]{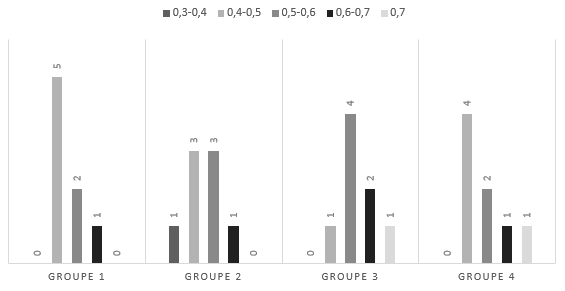}
		\caption{Participants par intervalle d'expertise sur les données de laboratoire.}
		\label{explab}
	\end{figure}

Les degrés d'expertise ont été calculés d'une part sur les données de laboratoire et d'autre part sur les notes issues des plateformes de {\em crowdsourcing}. Pour les données du laboratoire, les résultats sont présentés par intervalles de degré d'expertise de longueur 0.1 sur la figure~\ref{explab}.
On a obtenu 31 personnes (sur 32) avec un degré d'expertise supérieur à 0.4 (seuil choisi en comparaison des données issues de la plateforme) montrant ainsi la fiabilité des réponses des personnes en laboratoire. L'intervalle $[0.4,0.5]$ contient le plus de personnes.

\begin{figure}[!h]
	\includegraphics[width=0.9\textwidth]{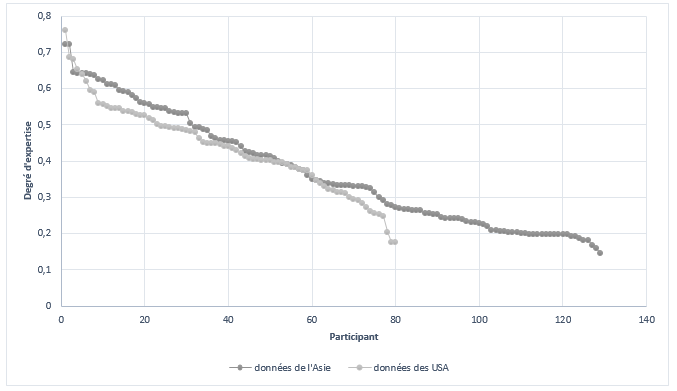}
	\caption{Distribution des données}
	\label{dasie}
\end{figure}

Dans un premier temps, les distributions des ensembles de tous les panels sur les données issues des plateformes de {\em crowdsourcing} sont représentées sur la figure \ref{dasie}. Sur ces distributions, nous notons un petit saut sur l'intervalle $[0.4,0.5]$ qui permet ainsi de déterminer le seuil d'expertise à prendre en compte pour discriminer les participants. 

Les degrés d'expertise varient dans un intervalle plus large que celui du laboratoire (de 0.1 à 0.7). Un premier facteur pouvant expliquer ces résultats serait le manque de sérieux chez un plus grand nombre de participants sur la plateforme de {\em crowdsourcing}. D'autre part, les conditions d'écoute (environnement sonore, casque ou haut-parleur(s), PC utilisé) sont variables d'un participant à l'autre, d'un {\em hit} à l'autre contrairement au laboratoire, et peuvent influer sur la qualité des réponses des participants. Dans ce travail, nous n'avons pas souhaité imposer des   conditions d'expérimentation afin de placer les travailleurs dans un contexte familier. De plus, en comparant les deux distributions, on remarque une petite différence entre les deux campagnes. Par exemple, l'intervalle d'expertise [0.1,0.2] est presque absent pour les données de la campagne américaine (2 personnes sur l'ensemble des panels {\em cf.} courbe grise) alors que pour l'Asie cet intervalle contient 19 participants ({\em cf.} courbe noire). D'autre part, pour l'Asie l'intervalle [0.2,0.3] contient le plus de participants alors que pour les USA, il s'agit de l'intervalle [0.4,0.5]. Les différences observées entre les deux campagnes peuvent s’expliquer à travers les différences culturelles entre les deux régions, et notamment une plus grande proximité culturelle avec les extraits musicaux choisis (musique occidentale) pour la campagne américaine.

Nous retenons comme seuils dans une première analyse, les seuils de $0.4$ et $0.5$ qui sont proches des sauts dans les distributions (\ref{dasie}). Les participants retenus sont ceux ayant un degré d'expertise supérieur au seuil considéré. La moyenne de leurs réponses sera prise en compte pour l'évaluation de la qualité audio. Nous comparons ainsi les données issues des deux campagnes sur les plateformes de {\em crowdsourcing} avec celles obtenues en laboratoire selon ces deux seuils.
	\begin{figure}[!h]
		\includegraphics[width=1\textwidth]{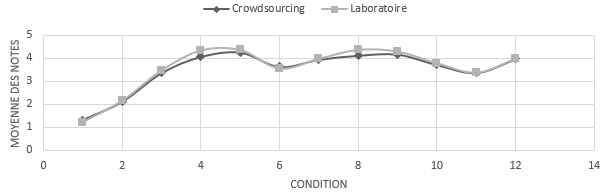}
		\caption{Comparaison des courbes laboratoire/crowdsourcing des données fusionnées pour un seuil d'expertise de 0.4.}
		\label{mbm2}

	\end{figure}
	\begin{figure}[!h]
		\includegraphics[width=1\textwidth]{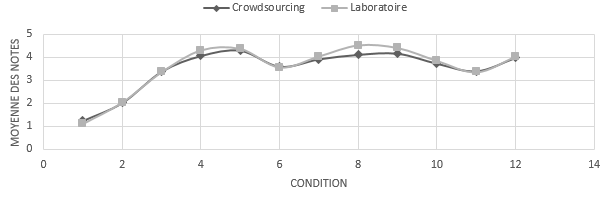}
		\caption{Comparaison des courbes laboratoire/crowdsourcing des données fusionnées pour un seuil d'expertise de 0.5.}
		\label{mbm4}
	\end{figure}
	
	Nous remarquons que les courbes obtenues (laboratoire et {\em crowdsourcing}) pour le premier seuil $0.4$ (\ref{mbm2}) sont plus proches les unes des autres que pour un seuil à 0.5. Néanmoins, les courbes obtenues pour le deuxième seuil $0.5$ (\ref{mbm4}) montrent que les participants qualifiés d'experts selon ce critère ont bien réussi à différencier les 5 premières conditions de référence (MNRU) de qualité connue.
	
	Idéalement, la courbe sur les cinq premiers morceaux devrait être une droite car elle correspond aux MNRUs. Cependant, ces résultats sont expliqués par des comportements habituels que l'on retrouve ici sur les données de laboratoire et des plateformes. La proximité des deux courbes montre l'intérêt de réaliser ce type d'évaluation sur des plateformes de {\em crowdsourcing}, une fois que les participants les plus experts aient été sélectionnés. 
	
\section{Conclusion et discussion}

Dans ce travail nous avons proposé une approche originale de calcul d'expertise pour des participants à une évaluation subjective de qualité audio à travers des tests d'écoute. L'approche proposée est fondée sur une modélisation des notations des participants sur la forme de graphe. Tenant compte de données dont l'ordre de préférence attendu est connu, nous avons développé une mesure de comparaison de deux graphes. Ainsi, l'approche est fondée sur quatre critères à partir desquels quatre fonctions de masse ont été définies afin de tenir compte des imperfections possibles des réponses des participants. A partir de ces fonctions de masse, un degré d'expertise est calculé pour chaque participant permettant ainsi de ne considérer que les participants ayant un degré d'expertise suffisant.

Les résultats comparant des données issues de deux campagnes de {\em crowdsourcing} et de laboratoire montrent l'intérêt de réaliser de telles évaluations à partir de plateformes de {\em crowdsourcing}. Il est cependant nécessaire d'évaluer correctement un degré d'expertise afin de ne pas considérer toutes les réponses issues des plateformes de {\em crowdsourcing}. L'approche développée dans ce travail pour l'évaluation des degrés d'expertise permet bien d'écarter les participants sans réponses pertinentes pour la tâche d'évaluation de la qualité audio.

\bibliographystyle{rnti}
\bibliography{refdis}

\providecommand\Fr{}
\providecommand\Eng{}
\providecommand\andname{and}
\providecommand\andnamec{and}

\begin{thebibliography}{}


\bibitem[{Ben~Rjab et~al.}(2016){Ben~Rjab, Kharoune, Miklos, \andnamec{}
  Martin}]{Am2}
Ben~Rjab, A., M.~Kharoune, Z.~Miklos, \andname{} A.~Martin (2016).
\newblock Characterization of experts in crowdsourcing platforms.
\newblock In {\em The 4th International Conference on Belief Functions}.

\bibitem[{Ben~Rjab et~al.}(2015){Ben~Rjab, Kharoune, Miklos, Martin,
  \andnamec{} Ben~Yaghlane}]{Am1}
Ben~Rjab, A., M.~Kharoune, Z.~Miklos, A.~Martin, \andname{} B.~Ben~Yaghlane
  (2015).
\newblock Caractérisation d’experts dans les plate-formes de crowdsourcing.
\newblock In {\em 24 ème Conférence sur la Logique Floue et ses
  Applications}.

\bibitem[{Dawid \andnamec{} Skene}(1979){Dawid \andnamec{} Skene}]{DandS}
Dawid, A.~P. \andname{} A.~M. Skene (1979).
\newblock Maximum likelihood estimation of observer error-rates using the em
  algorithm.
\newblock {\em Journal of the Royal Statistical Society\/}~{\em 28\/}(1),
  20--28.

\bibitem[{Dempster}(1967){Dempster}]{dempster1967upper}
Dempster, A.~P. (1967).
\newblock Upper and lower probabilities induced by a multivalued mapping.
\newblock {\em The annals of mathematical statistics\/}, 325--339.

\bibitem[{Essaid et~al.}(2014){Essaid, Martin, Smits, \andnamec{}
  Yaghlane}]{Essaid2014}
Essaid, A., A.~Martin, G.~Smits, \andname{} B.~B. Yaghlane (2014).
\newblock A distance-based decision in the credal level.
\newblock In {\em Artificial Intelligence and Symbolic Computation - 12th
  International Conference, {AISC} 2014, Seville, Spain, December 11-13, 2014.
  Proceedings}, pp.\  147--156.

\bibitem[{Howe}(2006){Howe}]{howe2006rise}
Howe, J. (2006).
\newblock The rise of crowdsourcing.
\newblock {\em Wired magazine\/}~{\em 14\/}(6), 1--4.

\bibitem[{Ipeirotis}(2010){Ipeirotis}]{panos}
Ipeirotis, P. (2010).
\newblock Worker evaluation in crowdsourcing: Gold data or multiple workers?

\bibitem[{Ipeirotis et~al.}(2010){Ipeirotis, Provost, \andnamec{} Wang}]{ipr}
Ipeirotis, P.~G., F.~Provost, \andname{} J.~Wang (2010).
\newblock Machine-learning for spammer detection in crowd-sourcing.
\newblock In {\em HCOMP '10 Proceedings of the ACM SIGKDD Workshop on Human
  Computation}.

\bibitem[{ITU}(1996){ITU}]{ITU96}
ITU (1996).
\newblock Modulated noise reference unit ({MNRU}).
\newblock Technical Report ITU-T P.810, International Telecommunication Union.

\bibitem[{Jouili}(2011){Jouili}]{Jouili11}
Jouili, S. (2011).
\newblock {\em Indexation de masses de documents graphiques : approches
  structurelles}.
\newblock Ph.\ D. thesis, Université Nancy II.

\bibitem[{Jousselme et~al.}(2001){Jousselme, Grenier, \andnamec{}
  Boss{\'e}}]{Jousselme}
Jousselme, A.-L., D.~Grenier, \andname{} {\'E}.~Boss{\'e} (2001).
\newblock A new distance between two bodies of evidence.
\newblock {\em Information fusion\/}~{\em 2\/}(2), 91--101.

\bibitem[{Le et~al.}(2010){Le, Edmonds, Hester, \andnamec{} Biewald}]{Le}
Le, J., A.~Edmonds, V.~Hester, \andname{} L.~Biewald (2010).
\newblock Ensuring quality in crowdsourced search relevance evaluation: The
  effects of training question distribution.
\newblock In {\em Workshop on Crowdsourcing for Search Evaluation}.

\bibitem[{Philips}(2011){Philips}]{patric}
Philips, P. (2011).
\newblock Enterprise crowdsourcing or: How i learned to stop worrying and trust
  the crowd.

\bibitem[{Raykar \andnamec{} Yu}(2012){Raykar \andnamec{} Yu}]{raykar2}
Raykar, V.~C. \andname{} S.~Yu (2012).
\newblock Annotation models for crowdsourced ordinal data.
\newblock {\em Journal of Machine Learning Research\/}~{\em 13}.

\bibitem[{Raykar et~al.}(2010){Raykar, Yu, Zhao, Hermosillo~Valadez, Florin,
  Bogoni, \andnamec{} Moy}]{raykar}
Raykar, V.~C., S.~Yu, L.~H. Zhao, G.~Hermosillo~Valadez, C.~Florin, L.~Bogoni,
  \andname{} L.~Moy (2010).
\newblock Learning from crowds.
\newblock {\em Journal of Machine Learning Research\/}~{\em 11}, 1297--1322.

\bibitem[{Shafer et~al.}(1976){Shafer et~al.}]{shafer1976mathematical}
Shafer, G. et~al. (1976).
\newblock {\em A mathematical theory of evidence}, Volume~1.
\newblock Princeton university press Princeton.

\bibitem[{Smets}(1990){Smets}]{smets}
Smets, P. (1990).
\newblock The combination of evidence in the transferable belief model.
\newblock ~{\em 12}, 447 -- 458.

\bibitem[{Smyth et~al.}(1995){Smyth, Fayyad, Burl, Perona, \andnamec{}
  Baldi}]{smyth}
Smyth, P., U.~Fayyad, M.~Burl, P.~Perona, \andname{} P.~Baldi (1995).
\newblock Inferring ground truth from subjective labelling of venus images.
\newblock {\em Advances in Neural Information Processing Systems\/}~{\em 7},
  1085--1092.

\end{thebibliography}
\Fr

\end{document}